\documentclass[sigconf]{acmart}
\AtBeginDocument{%
  \providecommand\BibTeX{{%
    \normalfont B\kern-0.5em{\scshape i\kern-0.25em b}\kern-0.8em\TeX}}}

\usepackage{color}
\usepackage{amsfonts}
\usepackage{bm}
\usepackage{amsmath}
\usepackage{amsthm}
\theoremstyle{definition}
\newtheorem{definition}{Definition}
\newtheorem{problem}{Problem}
\usepackage{multirow}
\usepackage{subfigure}
\usepackage{algorithm}
\usepackage{algorithmic}

\newcommand{\mname}{\texttt{StageNet}}

\copyrightyear{2020}
\acmYear{2020}
\setcopyright{iw3c2w3}
\acmConference[WWW '20]{Proceedings of The Web Conference 2020}{April 20--24, 2020}{Taipei, Taiwan}
\acmBooktitle{Proceedings of The Web Conference 2020 (WWW '20), April 20--24, 2020, Taipei, Taiwan}
\acmPrice{}
\acmDOI{10.1145/3366423.3380136}
\acmISBN{978-1-4503-7023-3/20/04}

\begin{document}

\title{\mname: Stage-Aware Neural Networks for Health Risk Prediction}

\author{Junyi Gao$^{1,2}$,  Cao Xiao$^3$, Yasha Wang*$^{1,2}$, Wen Tang$^4$, Lucas M. Glass$^{3,5}$,  Jimeng Sun$^{6,7}$}

\thanks{* corresponding author.}

\affiliation{Key Laboratory of High Confidence Software Technologies, Ministry of Education of China$^1$, National Engineering Research Center of Software Engineering, Peking University$^2$, IQVIA$^3$, Department of Nephrology, Peking University Third Hospital$^4$, Temple University$^5$, University of Illinois Urbana-Champaign$^6$, Georgia Institute of Technology$^7$}

\renewcommand{\shortauthors}{Gao, et al.}

\begin{abstract}
Deep learning has demonstrated success in health risk prediction especially for patients with chronic and progressing conditions. Most existing works focus on learning disease patterns from longitudinal patient data, but pay little attention to the disease progression stage itself. To fill the gap, we propose a \underline{Stage}-aware neural \underline{Net}work (\mname) model to extract disease stage information from patient data and integrate it into risk prediction.
\mname \  is enabled by (1) a stage-aware long short-term memory (LSTM) module that extracts health stage variations unsupervisedly; (2) a stage-adaptive convolutional module that incorporates stage-related progression patterns into risk prediction. We evaluate \mname \  on two real-world datasets and show that \mname \  outperforms state-of-the-art models in risk prediction task and patient subtyping task. Compared to the best baseline model, \mname \  achieves up to 12\% higher AUPRC for risk prediction task on two real-world patient datasets. \mname \  also achieves over 58\% higher Calinski-Harabasz score (a cluster quality metric) for a patient subtyping task.
\end{abstract}



\maketitle

\section{Introduction}

Deep learning has demonstrated early successes in health risk prediction using electronic health records (EHR) data especially for patients with chronic and progressing conditions such as heart diseases and Parkinson's disease
\cite{ma2018health,choi2016retain,heo2018uncertainty,baytas2017patient}. Most existing works focus on the extraction of disease patterns by modeling the relationship between disease progression and time from longitudinal patient data. For example, Pham {\it et al.}~\cite{pham2016deepcare} utilized RNN and a multiscale pooling layer to integrate temporal disease patterns from different time scales. Baytas {\it et al.}~\cite{baytas2017patient} and Ma {\it et al.}~\cite{ma2018health} simulates progression of patients' status by using temporal information to decay the information from historical timesteps.  


Despite these successes, the aforementioned works~\cite{ma2018health,baytas2017patient,pham2016deepcare,zheng2017capturing} implicitly assume that disease progression is smooth in time --- the longer the time is, the greater the change of health status will be. However, in reality, disease progression speed can vary significantly depending on the underlying {\it disease stage}. 

\begin{figure}[h!]
\vskip -1em
  \centering
  \subfigure[Albumin]{
      \includegraphics[scale=0.4]{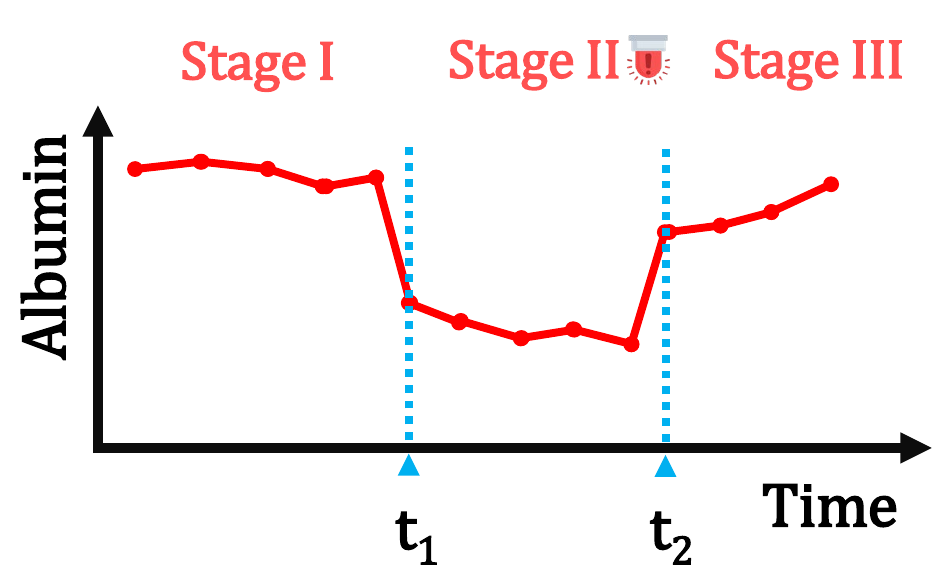}
      \label{fig:albumin}
  }
  \subfigure[hs-CRP]{
      \includegraphics[scale=0.42]{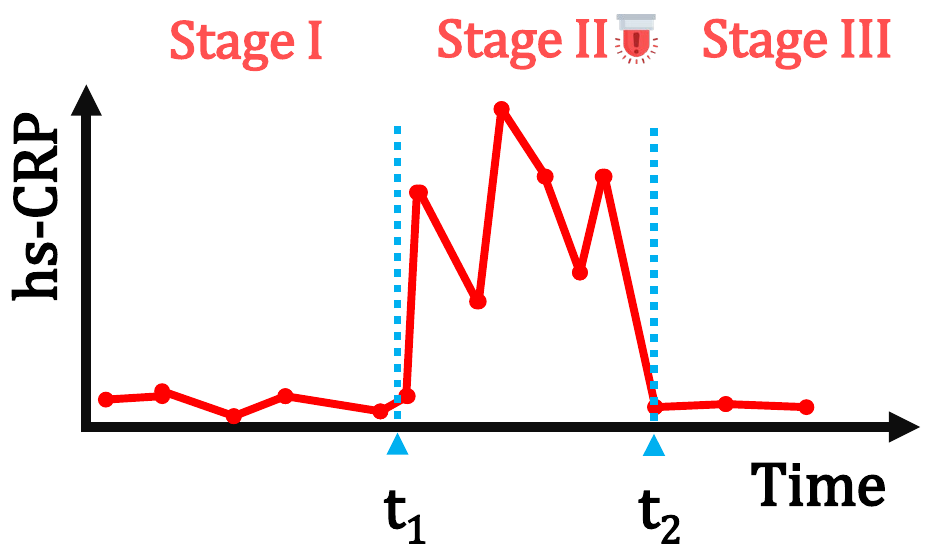}
      \label{fig:hscrp}
  }
  \caption{The change of biomarkers during different stages of an ESRD patient}
  \label{fig:motivating}
\vskip -1em
\end{figure}

\noindent{\bf Motivating example:}  Fig.~\ref{fig:motivating} plots the variation of albumin and hypersensitive C-reactive protein (hs-CRP) of an end-stage renal disease (ESRD) patient. We can observe the patient's health status between $t_{1}$ and $t_{2}$ is different from the rest time period. The sudden decline of albumin and fluctuation of hs-CRP indicate the patient's status is deteriorating rapidly. We consider the patient's health condition enters a new stage at $t_{1}$. 
The disease stage here refers to a time period with consistent status progression. It is not specific to a single disease (e.g. Alzheimer's Stage 1 or 2) but considers generally all comorbidities a patient has. For example, deteriorating and recovering are two different stages.

\noindent{\bf Challenges:} Disease progression stages indicate different risk levels. However, we still have the following challenges to utilize stage information in risk prediction tasks.
\begin{enumerate}
\item \textbf{How to extract disease progression stages from complex EHR data?}
Although important disease stage information is often unavailable in data and sometimes not even clearly defined for many diseases. Many temporal models such as LSTM have gating mechanism to control what historical information to remember or forget. However, the explicit underlying stage or change point is not clearly specified by those models. 

\item \textbf{How to leverage disease progression stage information for more accurate risk prediction?} Intuitively disease progression patterns should be similar within one stage but different across stages, as shown in Fig.~\ref{fig:hscrp}. Probably due to the absence of stage information in the data, disease stage information is largely ignored in predictive modeling of EHR data.  
Since the progression patterns relate to health risks~\cite{halm1998time,plochl1996nutritional}, the model should learn to extract and select informative patterns from each disease stage for more accurate risk prediction.
\end{enumerate}

\noindent{\bf Contributions:}
To address these challenges, we propose a new \underline{Stage}-aware neural \underline{Net}work model (\mname) to extract disease stage information from patient data and integrate it into risk prediction. \mname \  consists of two modules, a stage-aware LSTM module and a stage-adaptive convolutional module. Our \mname \  model is enabled by the following technical contributions.

\begin{enumerate}
    \item  \textbf{Extract disease progression stage without supervision}. The stage-aware LSTM module of \mname \  can capture the stage variation of patients' health conditions unsupervisedly. Specifically, we integrate inter-visit time information into LSTM cell states, which enables each dimension in the cell state to decide the storage proportion between long-term progression and short-term health status information. With such a design, \mname \  can evaluate the variation of patient health conditions compared with previous stages.
    \item \textbf{Learn disease progression patterns from disease stage information}. \mname \  further incorporates extracted disease progression stage information into the convolution operation, which can learn progression patterns closely related to the current stage. We further re-calibrate these patterns to emphasize informative patterns for outcome prediction.
\end{enumerate}

We evaluated \mname \  with both health risk prediction task and patient subtyping task on real-world urgent care MIMIC-III dataset and end-stage renal disease dataset. Risk prediction results show that \mname \  consistently outperforms all state-of-the-art models on both datasets in terms of different evaluation metrics. The improvement of \mname \  is up to 12\% in AUPRC compared to the best baseline model. The patient subtyping results show that \mname \  performs better than baseline models in identifying discriminative patient subgroups.

\section{Related Work}
\noindent\textbf{Disease Progression Modeling} 
Recent years, various deep learning models have been proposed to model disease progression. One solution is to model the disease trajectory using Markov-based models. For example, Sukkar \textit{et al.}~\cite{sukkar2012disease} applied Hidden Markov Model (HMM) to learn Alzheimer’s disease progression. 
Wang \textit{et al.}~\cite{wang2014unsupervised} proposed a probabilistic model to learn a continuous-time disease progression. Liu et al.~\cite{NIPS2015_5792} developed a continuous time HMM for modeling glaucoma and Alzheimer patients. However, these works assume Markov property which might not be true in practice when existing long term dependency.

Instead of modeling state transition probability, another line of efforts, to which our approach belongs, focus on utilizing patients' general health status progression to conduct clinical prediction tasks such as subtyping~\cite{baytas2017patient,galagali2018patient} or risk prediction~\cite{zheng2017capturing,pham2017predicting}. For example, T-LSTM~\cite{baytas2017patient} incorporated the elapsed time information into the standard LSTM architecture to model status progression in the presence of time irregularities. Health-ATM~\cite{ma2018health} used a time-aware convolutional layer by integrating time stamps into the original convolutional layer to model status progression. However, none of these works model disease progression from the stage perspective. To the best of our knowledge, we are the first work to explicitly extract disease stage progression information and utilize it in risk prediction tasks.

\noindent\textbf{Attention mechanism for feature re-calibration}
Feature re-calibration refers to using attention mechanism to emphasize informative input features or convolutional feature maps and suppress less informative ones. Feature re-calibration mechanism has achieved success in computer vision tasks. Hu {\it et al.}~\cite{hu2018squeeze} proposed Squeeze-and-Excitation Block to explicitly model the dependencies between the channels of convolutional features by aggregating convolutional feature maps across spatial dimensions. In EHR analysis, feature re-calibration mechanism is mainly used to provide model interpretability. For example, Choi \textit{et al.} proposed RETAIN to provide feature interpretability using attention over input features.

In this work, we re-calibrate disease progression patterns via extracting the progression theme at the current stage. Our model can adaptively emphasize most indicative features to help better predict patients' health risks at different stages.

\noindent{\bf Other related work:}
The underlying model of the stage-aware LSTM module in our work is related to the ordered neuron mechanism in \cite{shen2018ordered}, which proposes to solve NLP parsing tasks. However, unlike word sequences, the time irregularity in EHR data has clinical meanings. In this work, the inter-visit time intervals are used to help LSTM summarize stages of patients' status progression. 

\section{\mname\ Method}
\subsection{Overview}
Below we define the data and task studied in this work and provide the list of notations used in \mname \  in Table.~\ref{tab:notations}.
\begin{definition}[\textbf{Patient Records}]
In longitudinal EHR data, each patient can be represented as a sequence of multivariate observations $\bm{P} = [\bm{v}_{1}, \bm{v}_{2}, ..., \bm{v}_{T}]$, where $T$ is the number of visits of the patient. Each visit record $\bm{v}_{t} \in \mathbb{R}^{N_{v}}$ is a concatenation of clinical features (i.e., lab tests and measurements) including multi-hot and numerical values, where $N_{v}$ is the number of medical features at each visit. To model the time irregularity of visits, we also use $\bm{T} = [\Delta_{1}, \Delta_{2},...,\Delta_{t}]$ to represent the elapsed time between two visits, $\Delta_{1}=0$, specifically.
\end{definition}

\begin{problem}[\textbf{Health Risk prediction}]
Given a patient's visit records $\bm{P}$, we define dynamic patient risk prediction as to predict the health risk $\hat{y}_{t} \in \{0,1\}$ at the $t$-th timestep, which indicates the target outcome of interest (e.g., mortality, decompensation).
\end{problem}

\begin{table}[h!]
    \centering
      \caption{Notation definition}
    \label{tab:notations}
    \resizebox{0.92\columnwidth}{!}{
    \begin{tabular}{l|l}
        \hline
        Notation & Definition \\
        \hline
         $\bm{P}$; $\bm{T}$ & patient record sequence; time interval sequence \\
         $\bm{v}_{t} \in \mathbb{R}^{N_{v}}$ & multivariate visit record at the $t$-th visit \\
         $\Delta_{t}$ & time interval between $\bm{v}_{t}$ and $\bm{v}_{t-1}$ \\
         \hline
         $y_{t}$ & ground truth of prediction targets at the $t$-th visit \\
         $\hat{y}_t$ & predictions at the $t$-th visit \\
         \hline
         $\bm{h}_{t} \in \mathbb{R}^{N_{h}}$ & hidden state of LSTM at the $t$-th timestep \\
         $\bm{c}_{t} \in \mathbb{R}^{N_{h}}$ & cell state of LSTM at the $t$-th timestep \\
         $\bm{u}_{t} \in \mathbb{R}^{N_{m}}$ & extracted progression patterns within the current stage \\
         $\bm{z}_{t} \in \mathbb{R}^{N_{h}}$ & status progression theme at the current stage \\
         $\bm{x}_{t} \in \mathbb{R}^{N_{m}}$ & re-calibration weights of progression patterns \\
         $\widetilde{\bm{u}}_{t} \in \mathbb{R}^{N_{m}}$ & re-calibrated stage-adaptive progression patterns \\
         \hline
         $s_{t}$ & stage variation at the $t$-th timestep \\
         $\Delta s^{t}_{i}$ & stage variation from $\bm{h}_{t-K+i}$ to $\bm{h}_{t}$ \\
         \hline
         $\widetilde{f}_{t}$;  $\widetilde{i}_{t}$ & master forget gate; master input gate \\
         \hline
    \end{tabular}}
\end{table}


\begin{figure}[h!]
\centering
\includegraphics[width=1\columnwidth]{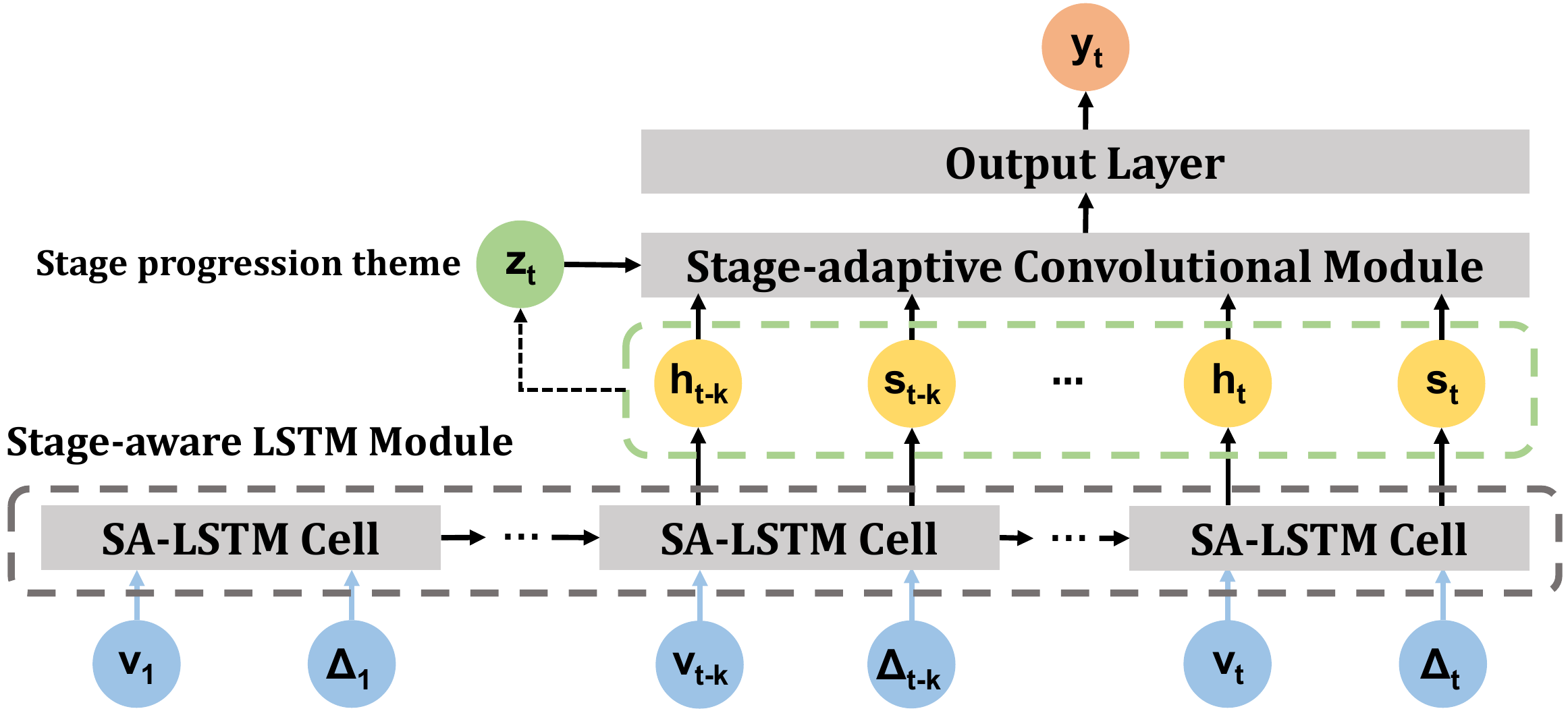}
 
\caption{\mname \  model: at the $t$-th timestep, the stage-aware LSTM takes current visit  $\bm{v}_{t}$ and elapsed time $\Delta_{t}$ as input to calculate current hidden state $\bm{h}_{t}$ and current stage variation $s_{t}$. Then the hidden states in the observation window $K$ will be fed into the stage-adaptive convolutional module. The convolutional module will extract progression patterns at the current stage and re-calibrate these patterns using the progression theme $\bm{z}_t$. Then the module will use re-calibrated patterns to predict health risk $\hat{y}_{t}$.}
 
\label{fig:model structure}
\end{figure}

As shown in Figure \ref{fig:model structure}, \mname \  comprises three modules: (1) a stage-aware LSTM module, (2) a stage-adaptive convolutional module, and (3) a prediction module. \mname \  takes patient EHR sequence $\bm{P}$ and time interval sequence $\bm{T}$ as input of the stage-aware LSTM module, and output the $t$-th time hidden state $\bm{h}_{t}$ along with stage variation factor $s_{t}$. 
Here $s_{t}$ represents the variation degree at the $t$-th visit compared to historical status -- a large $s_{t}$ indicating a higher chance of entering a new stage. 
We feed $\bm{h}_{t}$ and $s_{t}$ to the stage-adaptive convolutional module to extract patients' status progression patterns within an observation window $K$. The length of observation window $K$ is similar to convolutional kernel size, which decides the timescale of extracted convolutional patterns. The stage information $s_{t}$ is integrated to convolution operations to extract disease progression patterns that closely related to the current stage. We re-calibrate these patterns to emphasize informative patterns and suppress less useful ones via extracting the disease progression theme $\bm{z}_t$ at the current stage.
Last, we predict the health risk $\hat{y}_{t}$ based on these improved patient representations.

\subsection{Stage-aware LSTM module} 
\noindent\textbf{LSTM background} Given a sequence of patient health records $\bm{P}$ and a sequence of inter-visit time intervals $\bm{T}$, our goal is to infer the stage variation of the patient's health status (i.e., $s_{t}$) while constructing current health status (i.e., $\bm{h}_{t}$). Original LSTM consists of forget gate $\bm{f}_{t}$, input gate $\bm{i}_{t}$ and output gate $\bm{o}_{t}$. At the $t$-th timestep, the original LSTM takes previous hidden state $\bm{h}_{t-1}$, cell state $\bm{c}_{t-1}$ and current visit $\bm{v}_{t}$ as input, and output current hidden state $\bm{h}_{t}$ and cell state $\bm{c}_{t}$.
However, cell state $\bm{c}_t$  does not differentiate where the historical information is from (whether is from recent past or long history). \mname \  will use two new gates to make different dimensions in the cell state indicate different time scales.


\noindent\textbf{Goal}  
The key objective is to differentiate historical information in the cell state. If we can differentiate recent and old history in $\bm{c}_t$, we can determine whether the change to $\bm{c}_t$ is due to old history or recent history. Then a change mainly due to the recent history means the underlying disease stage has just changed. Hence, we can derive the disease stage change based on the change to the cell state. 

\noindent{\bf Idea} The idea is to make each dimension in the cell state represent patients’ status at a different time scale. Intuitively, the dimensions in cell state $\bm{c}_{t}$ can be divided into the low-ranking part (the first half of $\bm{c}_t$) and the high-ranking part (the second half of $\bm{c}_t$). The low-ranking part contains short-term health status information that only recent visits. And high-ranking part contains patients' long-term progression information that will last several visits or even the entire visit sequence. Note that because the low-ranking part is related to most recent visits, the update frequency of low-ranking dimensions is always higher than high-ranking dimensions. 
One simple way to enforce low and high ranking parts is to use two separate binary mask vectors. The low-ranking mask vector can be $[1,\ldots,1,0,\ldots, 0]$, while the high-ranking mask will be $[0,\ldots, 0, 1,\ldots, 1]$.

\noindent{\bf Soft masking} Instead of hard binary masks, we learn two soft mask vectors:
\begin{itemize}
    \item $\widetilde{\bm{f}_{t}}$ for high-ranking part (representing old history);
    \item $\widetilde{\bm{i}_{t}}$ for low-ranking part (representing recent history).
\end{itemize}
In particular, the ranking is dynamically determined using information from patients' current visit $\bm{v}_{t}$ and historical health status $\bm{h}_{t-1}$, adjusted by the elapsed time $\Delta_{t}$ between two visits. We utilize and further extend the master forget gate $\widetilde{\bm{f}_{t}}$ and master input gate $\widetilde{\bm{i}_{t}}$ in \cite{shen2018ordered} by using time interval information as follows:
\begin{equation}
\begin{aligned}
        &\bm{p}_{\widetilde{f}} = softmax(\mathbf{W}_{\widetilde{f}}(\bm{v}_{t}\oplus\Delta_{t}) + \mathbf{U}_{\widetilde{f}}(\bm{h}_{t-1}\oplus\Delta_{t}) + \mathbf{b}_{\widetilde{f}}) \\
        &\bm{p}_{\widetilde{i}} = softmax(\mathbf{W}_{\widetilde{i}}(\bm{v}_{t}\oplus\Delta_{t}) + \mathbf{U}_{\widetilde{i}}(\bm{h}_{t-1}\oplus\Delta_{t}) + \mathbf{b}_{\widetilde{i}}) \\
        &\widetilde{\bm{f}_{t}} = \stackrel{\rightarrow}{\mathrm{cm}}(\bm{p}_{\widetilde{f}}) \\
        &\widetilde{\bm{i}_{t}} = \stackrel{\leftarrow}{\mathrm{cm}}(\bm{p}_{\widetilde{f}})
\end{aligned}
\label{Eq.master gate}
\end{equation}
where $\mathbf{b}$ is the bias, $\mathrm{cm}$ denotes the cumulative sum, the arrow above $\mathrm{cm}$ indicates the direction of cumulative sum and $\oplus$ denotes the concatenation operation. 
$\bm{p}_{\widetilde{f}}$ and $\bm{p}_{\widetilde{i}}$ correspond to the probabilistic distribution of dimensions of $\bm{c}_t$ for high-ranking (old history) and low-ranking (recent history), respectively. 
Following the properties of the $cm$ operation, the values in $\widetilde{\bm{f}_{t}}$ are monotonically increasing from 0 to 1 (e.g. $\{0.2,0.4,0.8,1\}$), and those in $\widetilde{\bm{i}_{t}}$ are monotonically decreasing from 1 to 0 (e.g. $\{1,0.8,0.4,0.2\}$). 
Hence, $\widetilde{\bm{f}_{t}}$  and $\widetilde{\bm{i}_{t}}$ can be used to serve as the soft mask for high-ranking part and low-ranking part, respectively. 

\begin{figure}[ht]
\centering
\includegraphics[width=0.9\columnwidth]{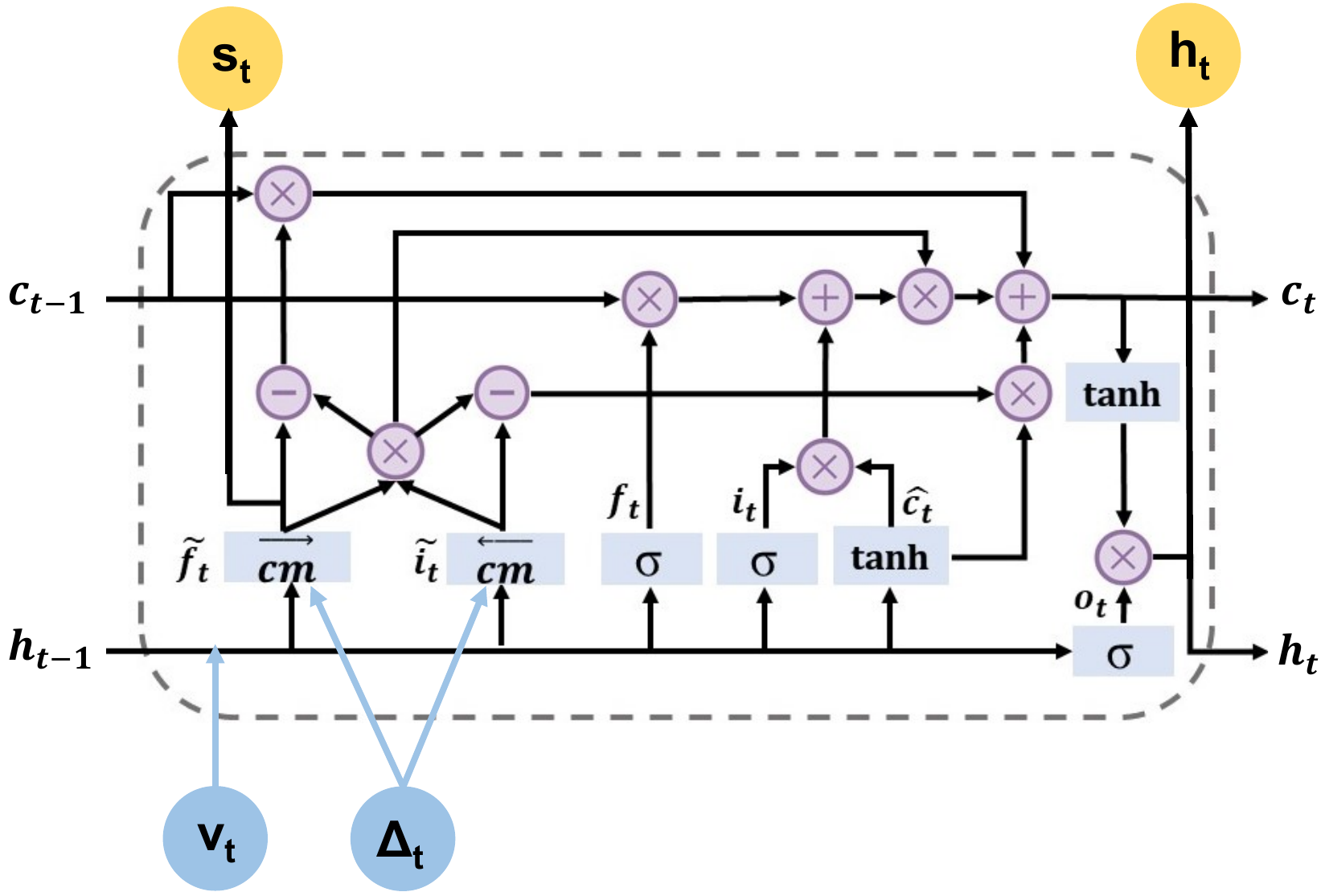}
\caption{The structure of stage-aware LSTM cell}
\label{fig:cell structure}
\end{figure}

\noindent{\bf Cell state update} We define the new calculation for current cell state $\bm{c}_{t}$ as: 
\begin{equation}
    \begin{split}
        \hat{\bm{c}}_{t} = &tanh(\bm{W}_{c}\bm{v}_{t}+\bm{U}_{c}\bm{h}_{t-1}+\bm{b}_{c}) \\
        \bm{w}_{t} = &\widetilde{\bm{f}_{t}} \odot \widetilde{\bm{i}_{t}} \\
        \bm{c}_{t} = &\bm{w}_{t} \odot (\bm{f}_{t} \odot \bm{c}_{t-1} + \bm{i}_{t} \odot \hat{\bm{c}}_{t}) \\
    &+(\widetilde{\bm{f}_{t}}-\bm{w}_{t})\odot \bm{c}_{t-1}+(\widetilde{\bm{i}_{t}}-\bm{w}_{t})\odot \hat{\bm{c}_{t}}\\
    \bm{h}_{t} = &\bm{o}_{t} \odot tanh(\bm{c}_{t}) \\
    \end{split}
\label{Eq.ct}
\end{equation}
where the calculation of intermediate cell state $\hat{\bm{c}_{t}}$ and hidden state $\bm{h}_{t}$ are same with the original LSTM. Values in $\widetilde{\bm{f}_{t}}$ are used to decide which dimensions in $\bm{c}_{t}$ to store long-term information about status progression (i.e. $\bm{c}_{t-1}$), and values in $\widetilde{\bm{i}_{t}}$ decide which dimensions to store short-term information (i.e. $\hat{\bm{c}}_{t}$). In Eq. \ref{Eq.ct}, $\bm{w}_{t}$ decide which dimensions to store the overlap part between $\bm{c}_{t-1}$ and $\hat{\bm{c}}_{t}$. Besides the overlapping information, the independent (non-overlapping) information in $\bm{c}_{t-1}$ and $\hat{\bm{c}}_{t}$ are stored into $c_t$ based on the values of $(\widetilde{\bm{f}_{t}}-\bm{w}_{t})$ and $(\widetilde{\bm{i}_{t}}-\bm{w}_{t})$, respectively. 
Fig~\ref{fig:cell structure} shows the structure of stage-aware LSTM.

\noindent{\bf Stage progression variation}
Since values in $\widetilde{\bm{f}_{t}}$ decide where to store progression information, we denote $s_{t}$ as:

\begin{equation}
    s_{t} = argmax(\bm{p}_{\widetilde{f}})
\end{equation}
 The value of $s_{t}$ decides how much history information is used to calculate the current $\bm{c}_{t}$. If $s_{t}$ is large, there is almost no historical state information in current cell state, which means that the patient's current health status have changed a lot compared to history status. In other word, a large $s_{t}$ may indicate that the patient's status may have entered a new stage.
 
Since $argmax$ function is non-differentiable, we use the following equation to estimate $s_{t}$:
\begin{equation}
    s_{t} \approx \sum_{i=1}^{N_{h}}{i\times \bm{p}_{\widetilde{f}}(i)} = N_{h}(1-\frac{1}{N_{h}}\sum_{i=1}^{N_{h}}{\widetilde{\bm{f}_{t}}(i)})+1
\end{equation}
where $\widetilde{\bm{f}_{t}}(i)$ and $\bm{p}_{\widetilde{f}}(i)$ are the $i$-th values in $\widetilde{\bm{f}_{t}}$ and $\bm{p}_{\widetilde{f}}$.

\noindent{\bf Example illustration} We use a toy example to better illustrate how $\widetilde{\bm{i}_{t}}$ and $\widetilde{\bm{f}_{t}}$ store patients' health status at different timescales and summarize stage variation factor $s_{t}$. Assume $N_{h}=5$, $\bm{p}_{\widetilde{i}}=[0,0,0,1,0]$ and $\bm{p}_{\widetilde{f}}=[0,0,1,0,0]$. According to Eq.~\ref{Eq.master gate} and \ref{Eq.ct}, $\widetilde{\bm{f}_{t}}=[0,0,1,1,1]$, $\widetilde{\bm{i}_{t}}=[1,1,1,1,0]$ and $\bm{w}_{t}=[0,0,1,1,0]$. 
And $s_{t}=2$. $s_{t}$ indicates the variation of patients' stage.

The $\bm{c}_{t}$ is calculated as:
\begin{equation}
\begin{split}
    \bm{c}_{t} = &[0,0,1,1,0] \odot (\bm{f}_{t} \odot \bm{c}_{t-1} + \bm{i}_{t} \odot \hat{\bm{c}}_{t}) \\
    &+[0,0,0,0,1]\odot \bm{c}_{t-1}+[1,1,0,0,0]\odot \hat{\bm{c}_{t}}
\end{split}
\end{equation}
The fifth dimension in $\bm{c}_{t}$ is used to store long-term progression information from $\bm{c}_{t-1}$, and the first and second dimensions are used to store short-term health status from $\hat{\bm{c}}_{t}$. The third and fourth dimensions are used to store overlapping information. However, because we use $softmax$ activation, the actual values in $\widetilde{\bm{f}_{t}}$ and $\widetilde{\bm{i}_{t}}$ are decimals instead of 0 and 1. 

As $\widetilde{\bm{f}_{t}}$ and $\widetilde{\bm{i}_{t}}$ only focus on coarse-grained control, in practice, we reduce the dimension of $\widetilde{\bm{f}_{t}}$ and $\widetilde{\bm{i}_{t}}$ to $N_{m} = N_{h}/C$ similar to \cite{shen2018ordered}, where $C$ is a chunk size factor. Therefore, every dimension within each $C$-sized chunk shares the same master gates. A smaller $C$ can make the model describe patients' status variation in more details.\\

\subsection{Stage-adaptive convolutional module}
To leverage the stage information learned from LSTM, we develop a stage-adaptive convolutional module on top of the recurrent layer to extract and re-calibrate patient health progression patterns for risk prediction.

\begin{figure}[ht]
\centering
\includegraphics[width=0.65\columnwidth]{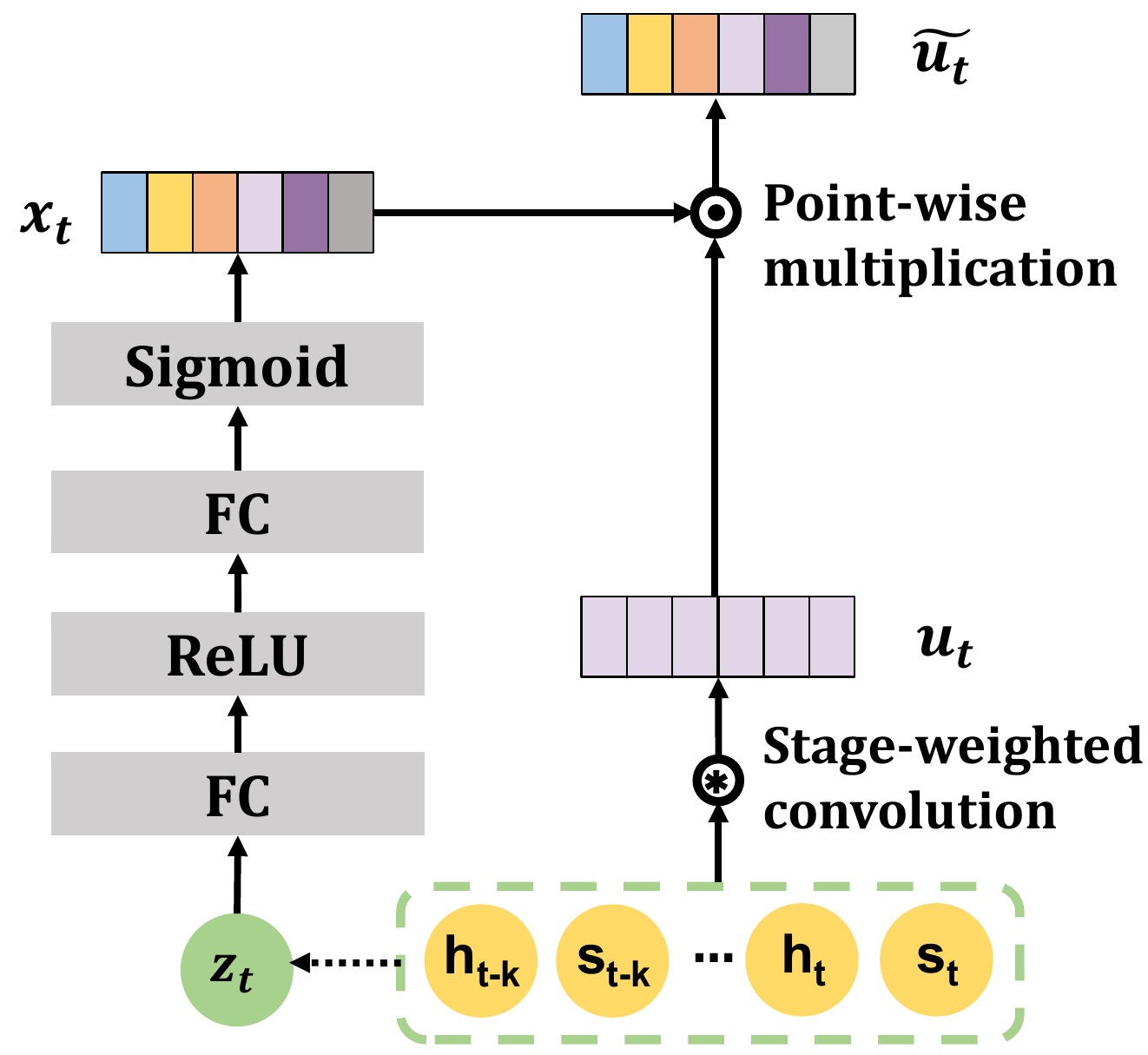}
\caption{Stage-adaptive convolutional module: The module takes historical hidden state $\bm{h}_{t}$ and stage variation $s_{t}$ within the observation window $K$ as input, and learn progression patterns $\bm{u}_{t}$ using stage-weighted convolution operation. These patterns are re-calibrated to emphasise the most informative patterns at the current stage via extracting the current progression theme $z_{t}$.}
\label{fig:recalibration}
\end{figure}

Progression patterns of patients' health status are critical in predicting patients' risks \cite{yeh1984validation,halm1998time}. Since there are many medical research indicate that these progression patterns are often similar within one stage, but vary across different stages~\cite{halm1998time,plochl1996nutritional,yeh1984validation}, we expect \mname \  can extract patterns that are closely related to patients' current stage using convolutional filters. We also design the model to adaptively select the most informative patterns for risk prediction at the current stage. We achieve this through three steps:
1) {\it learning stage progression patterns} 2) {\it extracting progression theme at the current stage} and 3) {\it re-calibrating progression patterns}. The structure of stage-adaptive convolutional module is shown in Fig.~\ref{fig:recalibration}.

\noindent{\bf Learning stage progression patterns:} We further extract the progression patterns of at the current stage by using stage-weighted convolution operators. We modify the original CNN by integrating disease stage information into the convolution operation. 

Mathematically, at the $t$-th time, we calculate the distance between stages of historical visits within the observation window and the stage of current visit as:
\begin{equation}
    \Delta \bm{s}^{t} = \mathrm{softmax}(\stackrel{\longrightarrow}{\mathrm{cm}}(s_{t-K},...,s_{t}))
\label{Eq.delta}
\end{equation}
where $K$ is the length of observation window. The length of observation window $K$ is similar to convolutional kernel size, which decides the timescale of extracted progression patterns. The values in $\Delta \bm{s}^{t}$ are monotonically decreasing from 1 to 0. We denote the $i$-th value in $\Delta \bm{s}^{t}$ as $\Delta \bm{s}^{t}_{i}$. A large $\Delta \bm{s}^{t}_{i}$ indicates that the stage of $\bm{h}_{t-K+i}$ is far from the current stage of $\bm{h}_{t}$.

In stage-weighted convolution operation at the $t$-th timestep, the convolutional module takes concatenated historical hidden states sequence of LSTM (i.e. $\bm{h}_{t-K:t} = [\bm{h}_{t-K},...,\bm{h}_{t}]$) as input. Different from the original convolutional layer, the weights of input variables are re-weighted by their stage distance $\Delta \bm{s}^{t}_{i}$ in each convolution computation as:
\begin{equation}
    \bm{u}_{t}^{i} = \bm{m}_{i} * \bm{h}_{t-K:t} = \sum_{j=1}^{N_{h}}\bm{m}_{i}^{j} * (\bm{h}_{t-K:t}^{j} \odot \Delta \bm{s}^{t})
\label{Eq.conv}
\end{equation}
where $*$ is the convolution operation, $\bm{m}_{c}=[\bm{m}_{c}^{1},...,\bm{m}_{c}^{N_{h}}]$, $\bm{u}_{t}^{i}\in \mathbb{R}^{K-S+1}$. $\bm{m}_{i}^{j}$ is a 1D convolution kernel representing a single channel of $\bm{m}_{i}$ that acts on the corresponding channel of $\bm{h}_{t-K:t}$ and $S$ is the kernel size. We use multiple kernels to generate extract different patterns, and the number of kernels is $N_{m}$. We concatenate the output of kernels to get the final convolution output as $\bm{u}_{t} = [\bm{u}_{t}^{1},...,\bm{u}_{t}^{N_{m}}]$. We set $S=K$ to make each kernel can extract progression patterns that represent the whole stage, so that the final dimension is $\bm{u}_{t}\in  \mathbb{R}^{N_{m}}$.

In Eq.~\ref{Eq.conv}, The weights of patients' historical health status are adjusted according to the distance of stages $\Delta \bm{s}^{t}$ in order to extract patterns that are closely related to the current stage. If the stage of a historical status is far from the current stage, it will have a lower weight in the stage-weighted convolution operation and vice versa. 

\noindent{\bf Extracting progression theme at the current stage:}
Since the output of the convolution operation $\bm{u}_{t}$ is the concatenation of multiple patterns extracted by different kernels. The importance of these patterns may vary and depends on patients' status at different stages. In order to select the most informative patterns at the current stage, we should provide the model with a global view of patients' status at the current stage as:
\begin{equation}
    \bm{z}_{t} = \frac{1}{K}\sum_{i=0}^{K}{\Delta s^{t}_{i}\bm{h}_{t-K+i}}
\label{Eq.global representation}
\end{equation}

The global status representation at the current stage $\bm{z}_{t}$ is the weighted average of hidden states within the observation window. $\bm{z}_{t}$ can be regarded as the {\it progression theme} of the current stage. The importance of different temporal patterns will be calculated based on this theme.

\noindent{\bf Re-calibrating progression patterns:}
After obtaining the status progression theme at the current stage, we map the reprensentation $\bm{z}_{t}$ to an importance vector $\bm{x}_{t}$, where the $i$-th value in $\bm{x}_{t}$ indicates the importance of the $i$-th extracted temporal pattern in $\bm{u}_{t}$. $\bm{x}_{t}$ is calculated as:
\begin{equation}
    \bm{x}_{t} = \sigma(\mathbf{W}_{x1} \delta (\mathbf{W}_{x2}\bm{z}_{t}))
\end{equation}
where $\sigma$ refers to the sigmoid functon, $\delta$ is the ReLU function, $\mathbf{W}_{x2} \in \mathbb{R}^{N_{x}\times N_{m}}$ and $\mathbf{W}_{x1} \in \mathbb{R}^{N_{m}\times N_{x}}$. We use two fully-connected (FC) layers to map the progression theme $\bm{z}_{t}$ to $\bm{x}_{t}$, i.e. a dimension-ality-reduction layer with ReLU activation to compress the representation while capturing the non-linearity, and a dimensionality-increasing layer to rescale the output to the original dimension of $\bm{u}_{t}$. We use sigmoid activation to generate the importance weights between 0 and 1.

Finally, the features in $\bm{u}_{t}$ is re-calibrated using $\bm{x}_{t}$ as:
\begin{equation}
  \widetilde{\bm{u}}_{t} = \bm{u}_{t} \odot \bm{x}_{t}
  \label{Eq.recalibrate ut}
\end{equation}

The re-calibration mechanism can be regarded as a channel-wise attention mechanism like \cite{hu2018squeeze}. However, the attention weights (i.e. the importance of each pattern) is calculated by patients' status progression theme at the current stage instead of using global average pooling to generate channel-wise statistics in \cite{hu2018squeeze} or calculating alignment between historical states.\\

\subsection{Prediction module}
The prediction layer takes the output of stage-adaptive convolutional module as input, and outputs a binary label $\hat{y}_{t}$, which indicates the patient's current health risk. Note that, we include residue connections between the convolutional module and the output layer. In order to achieve this, we set $N_{m} = N_{h}$. We compute $\hat{y}_{t}$ as:
\begin{equation}
    \hat{y}_{t} = \sigma(\mathbf{W}_{y}(\widetilde{\bm{u}}_{t}+\bm{h}_{t})+\mathbf{b}_{y})
\label{Eq.output}
\end{equation}
where $\mathbf{W}_{y} \in \mathbb{R}^{N_{h}}$. We choose the cross-entropy function to calculate the loss for each patient as: 
\begin{equation}
    \mathcal{L} = -\frac{1}{T}\sum_{t=1}^{T}{(y_{t}^{\top}\log(\hat{y}_{t})+(1-y_{t})^{\top}\log(1-\hat{y}_{t}))}
\label{Eq.loss}
\end{equation}
We use the Adam algorithm \cite{kingma2014adam} for optimization. We summarize \mname \  algorithm below.

\begin{algorithm}[htb] 
\caption{The \mname \  model} 
\label{alg:Framwork} 
\begin{algorithmic}
\REQUIRE ~~\\ 
Patient records $\bm{P} = [\bm{v}_{1}, \bm{v}_{2}, ..., \bm{v}_{T}]$, time interval sequence $\bm{T} = [\Delta_{1}, \Delta_{2},...,\Delta_{t}]$ and the length of observation window $K$. \\
\ENSURE ~~\\ 
Initialize $\bm{h}_{0}$ and $\bm{c}_{0}$ to zero.
\FOR{$i=1$ to $T$}
\STATE Input last $\bm{h}_{i-1}$ and $\bm{c}_{i-1}$ to stage-aware LSTM cell; \\ 
\STATE Obtain $\bm{h}_{i}$, $\bm{c}_{i}$ and $s_{i}$ via Eq. \ref{Eq.master gate} and \ref{Eq.ct}; \\
\STATE Compute $\Delta^{i}$ via Eq. \ref{Eq.delta}; \\
\STATE Compute progression patterns $\bm{u}_{i}$ via Eq.~\ref{Eq.conv}; \\
\STATE Calculate current stage progression theme $\bm{z}_{i}$ via Eq.~\ref{Eq.global representation}; \\
\STATE Compute re-calibrated patterns $\widetilde{\bm{u}}_{i}$ using Eq.~\ref{Eq.recalibrate ut}; \\
\STATE Compute current predicted risk $\hat{y}_{i}$ via Eq. \ref{Eq.output}; \\
\ENDFOR \\
Update the model's parameters by optimizing the loss in Eq. \ref{Eq.loss}.
\end{algorithmic}
\end{algorithm}

\section{Experiment}
We evaluated \mname \  model by comparing against other baselines on public dataset MIMIC-III and ESRD (i.e., end-stage renal disease) dataset. The code is provided in \footnote{https://github.com/v1xerunt/StageNet}.

\subsection{Dataset description}
We use the following data to evaluate our model.
\begin{itemize}
\item
\textbf{MIMIC-III Dataset} We use Intensive Care Unit (ICU) data from the publicly available Medical Information Mart for Intensive Care (MIMIC-III) database~\cite{johnson2016mimic}. Following the work \cite{harutyunyan2017multitask}, the cohort of 33, 678 unique patients with a total of 2,202,114 samples (i.e., records) is used. The raw data includes 17 physiologic variables at each visit, which is transformed into a 76-dimensional vector including numerical and one-hot encoded categorical clinical features. 

\item 
\textbf{End-Stage Renal Disease (ESRD) Dataset} 
We perform the mortality risk prediction on an end-stage renal disease dataset. There are many people suffered from ESRD in the world \cite{tangri2011determining, isakova2011fibroblast}. They face severe life threat and need lifelong treatment with periodic visits to the hospitals for multifarious tests (e.g., blood routine examination). The whole procedure needs a dynamic risk prediction system to help patients prevent adverse outcomes, based on the medical records collected along with the visits. The cleaned dataset consists of 656 patients and 13,091 visit records and the percentage of positive labels is 17.5\%. The raw data includes 17 numeric physiologic variables at each timestep. During and after data collection and analysis, the authors could not identify individual participants as patients' names were replaced by ID. We use patients' previous records to fill the missing data in order to prevent the leakage of future information.
\end{itemize}

\begin{table*}[h]
 
\caption{Performance comparison on MIMIC-III and ESRD dataset}
\begin{center} 
   
\begin{tabular}{llcccccc}
\toprule[0.7pt]
& & \multicolumn{3}{c}{\textbf{MIMIC-III}} & \multicolumn{3}{c}{\textbf{ESRD}} \\
&\textbf{Model}&\textbf{AUPRC}&\textbf{AUROC}&\textbf{min(Re, P+)}&\textbf{AUPRC}&\textbf{AUROC}&\textbf{min(Re, P+)} \\
\hline
\multirow{4}{*}{Baseline}&LSTM  & 0.280 (0.003) & 0.897 (0.002) & 0.324 (0.003) & 0.270 (0.029) & 0.805 (0.026) & 0.318 (0.015) \\
&ON-LSTM & 0.304 (0.002) & 0.895 (0.003) & 0.343 (0.004) & 0.291 (0.021) & 0.810 (0.021) & 0.333 (0.034) \\
&T-LSTM & 0.282 (0.004) & 0.895 (0.002) & 0.322 (0.005) & 0.276 (0.027) & 0.812 (0.026) & 0.331 (0.031) \\
&Decay-LSTM & 0.294 (0.002) & 0.893 (0.003) & 0.330 (0.004) & 0.289 (0.020) & 0.808 (0.022) & 0.328 (0.021) \\
&$\text{Health-ATM}^{-}$ & 0.291 (0.002) & 0.897 (0.003) & 0.325 (0.003) & 0.287 (0.021) & 0.810 (0.039) & 0.331 (0.025) \\
\hline
Reduced  & \mname-I & 0.313 (0.003) & 0.899 (0.003) & 0.360 (0.002) & 0.296 (0.014) & 0.814 (0.031) & 0.333 (0.018) \\
Model & \mname-II & 0.311 (0.003) &0.897 (0.002) & 0.358 (0.003) & 0.302 (0.029) & 0.812 (0.027) & 0.334 (0.017) \\
\hline
Proposed &\textbf{\mname} & \textbf{0.323 (0.002)} & \textbf{0.903 (0.002)} & \textbf{0.372 (0.003)} & \textbf{0.327 (0.022)} & \textbf{0.821 (0.024)} & \textbf{0.352 (0.019)} \\
\bottomrule[0.7pt]
\end{tabular}
\end{center}
\label{tab:result}
\end{table*}

\subsection{Baselines}
We evaluated \mname \  against the following baselines, which share some of the similar insights with \mname. It is worth noting that there are lots of state-of-the-art clinical prediction models which utilize attention mechanism to extract long-term dependencies in patients' historical visits \cite{lee2018diagnosis,choi2016retain,song2018attend}. However, their contribution is orthogonal to ours. We focus on capturing and utilizing stage information of patients' health status in EHR data. Our model \mname \  can be easily combined with attention mechanism.

\begin{itemize}
\item \textbf{LSTM} \cite{gers1999learning} The visit input at the $t$-th timestep is fed into the LSTM model. Then it directly output the prediction results based on the hidden state vector $\textbf{h}_{t}$.

\item \textbf{ON-LSTM} \cite{shen2018ordered} uses LSTM to model tree-like structures for natural language sequences by separately allocating hidden state dimensions with long and short-term information.

\item \textbf{T-LSTM} \cite{baytas2017patient} handles visit-level irregular time intervals by enabling time decay inside RNN cell, which makes older information less important. The original T-LSTM model is used for unsupervised clustering, and we modify it into a supervised learning model.

\item \textbf{Decay-LSTM} \cite{zheng2017capturing} uses feature-level time intervals to enable memory decay similar to T-LSTM. We adopt the decay mechanism on the input gate of LSTM. Decay-LSTM requires to input time intervals of each feature, and we also input this information to all the other models without loss of fairness.

\item $\textbf{Health-ATM}^{-}$ \cite{ma2018health} uses irregular time intervals to decay the information from historical timestep via a hybrid CRNN structure. The original model utilizes the target-aware attention mechanism to achieve disease prediction. Since our task doesn't have a specific target embedding to guide the attention, we remove the target-aware attention mechanism from Health-ATM.
\end{itemize}

We also compare \mname \  against its reduced models:

\begin{itemize}
\item \textbf{\mname-I} consists of regular LSTM and the stage-adaptive convolutional module. The weighted convolution operation is also replaced by regular convolution operation. We use the average of $\bm{h}_{t}$ within the observation window to calculate $\bm{z}_{t}$.

\item \textbf{\mname-II} only has stage-aware LSTM. The visit input at the $t$-th timestep is fed into the stage-aware LSTM model. Then it directly outputs the prediction results based on the hidden state vector $\textbf{h}_{t}$.
\end{itemize}

\subsection{Health Risk Prediction}
In this section, we report experimental results for following supervised tasks on two datasets. 

\begin{itemize}
\item
\textbf{Decompensation risk prediction} We perform the physiologic decompensation prediction task on MIMIC-III dataset. This task involves the detection of patients who are physiologically decompensating, which means conditions are deteriorating rapidly. Detection of decompensation is closely related to problems like condition monitoring and sepsis detection that have received significant attention from the machine learning community. The task is formulated as a binary classification task for predicting whether the patient's date of death (DOD) falls within the next 24 hours of the current time point. These labels are assigned to each hour, starting at four hours after admission to the ICU and ending when the patient dies or is discharged.

We truncate the length of samples to a reasonable limit (i.e. 400). We fix a test set of $15\%$ of patients, and divide the rest of the dataset into the training set and validation set with a proportion of 85\%:15\%. We fix the best model on the validation set and report the performance in the test set. We also report the standard deviation of the performance measures by bootstrapping the results on the test set for 10,000 times.

\item
\textbf{Mortality risk prediction} We perform the mortality risk prediction task on the ESRD dataset. Similar to the decompensation task, the mortality risk prediction task is formulated as a binary classification task to predict whether the patient will die within 12 months, and the predictions are made at each timestep. We evaluate the models with 10-fold cross-validation strategy and report the average performance and standard deviations.
\end{itemize}

\noindent\textbf{Implementation Details}
We train each model for 50 epochs on MIMIC-III dataset and 200 epochs on the ESRD dataset. The learning rate is set to 0.001. All methods are implemented in PyTorch 1.1 \cite{paszke2017automatic} and trained on a server equipped with an Intel Xeon E5-2620 Octa-Core CPU, 256GB Memory and a Titan V GPU. \\

\noindent\textbf{Evaluation Metrics}
Following existing works\cite{song2018attend,ma2018health,harutyunyan2017multitask}, we assess performance using area under the receiver operating characteristic curve (AUROC), area under the precision-recall curve (AUPRC), and the minimum of precision and sensitivity Min(Re,P+). The Min(Re,P+) is calculated as the maximum of min(recall, precision) on the precision-recall curve.
\subsubsection{Results}

Table \ref{tab:result} compares the performance of all models on both datasets. \mname \  consistently outperforms all state-of-the-art models on both datasets. On MIMIC-III dataset, \mname \  achieves 10\% higher AUPRC and min(Re,P+) compared to the best baseline model ON-LSTM and Health-ATM. On the ESRD dataset, \mname \  achieves 12\% higher AUPRC and 6\% higher min(Re, P+) compared to the best baseline model ON-LSTM. 

The reduced models \mname-I and \mname-II still outperform all state-of-the-art models in most cases on both datasets. It proves that extracting higher-level temporal variation features and summarizing stage information are both helpful for predicting patients' health risks. Among all baselines, ON-LSTM and Health-ATM achieve better performance in most cases due to handling the aforementioned challenges to some extent.



 


 


\subsubsection{Observations}
\begin{figure}[ht]
\centering
\includegraphics[width=0.8\columnwidth]{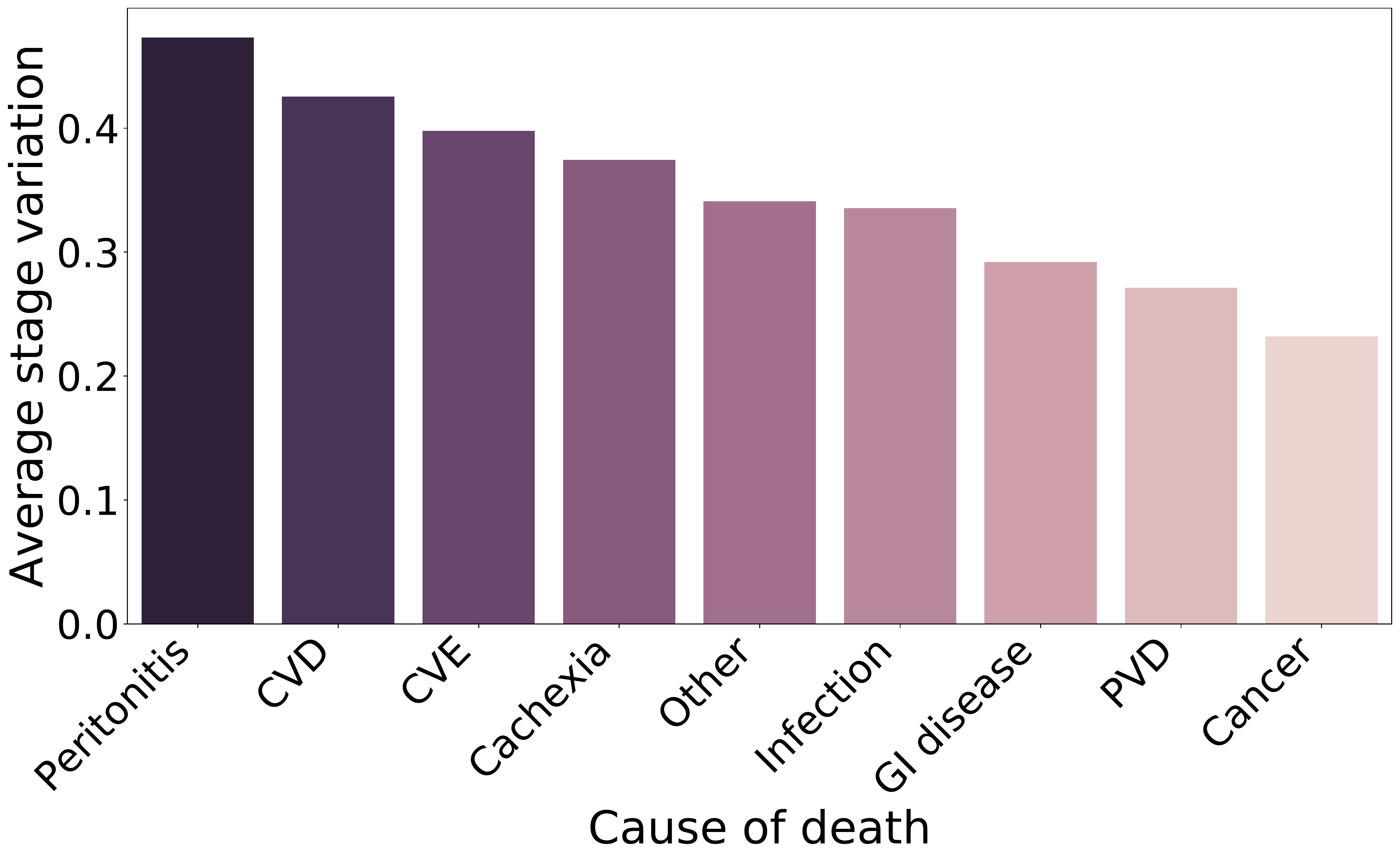}
 
\caption{Causes of death and status stability}
\label{fig:death_reason}
\end{figure}

~\\
\noindent\textbf{Health status stability vs. Cause of death}
In order to understand how different clinical events affect disease progression, we further analyze the health status stability of patients with different causes of death in the ESRD dataset. At each timestep, \mname \  will output a scalar $s_{t}$, which indicates the variation of patients' stage. A large $s_{t}$ indicates the patient's current health status has changed a lot compared to history status (i.e. enter a new stage). We compute the average stage variation (i.e. the average of $s_{t}$) for each patient. A patient with stable health status will have a low $s_{t}$. The average stage variation of patients with different causes of death is shown in Fig.~\ref{fig:death_reason}.

The result shows that patients who died of peritonitis, cardiovascular (CVD) and cerebrovascular (CVE) have the highest $s_{t}$, which means that their health status are very unstable. These diseases are acute diseases and have high mortality rate~\cite{kannel1987heart,fried1996peritonitis,estanol1975cardiac}. The health status of patients who have these diseases tends to deteriorate rapidly in a short period of time, which explains why our model believes these patients' status are highly unstable. In contrast, patients with cancer have the most stable status compared to other patients, since their health status often deteriorates more chronically and have longer survival time compared to the patients with acute cardiac diseases~\cite{derogatis1979psychological,prentice1978regression}. Clinicians should pay more attention to patients with heart disease or peritonitis history in order to take timely interventions.\\

\noindent\textbf{Health status stability vs. Health risk}.
The stability of patients' health status is an important indicator to evaluate patients' health risk~\cite{plochl1996nutritional,halm1998time,yeh1984validation}. At each visit, our model will evaluate the patient's current health status and output a health risk score. We divide patients' visits into three groups according to predicted health risk: Low risk (risk score <= 0.4), Medium risk (0.4 < risk score <= 0.7) and High risk (risk score >= 0.7). We compute the average stage variation of each group to explore Table ~\ref{tab:stage_risk}.

\begin{table}[h]
\centering
  \caption{Health risk levels and status stability}
   
  \label{tab:stage_risk}
  \resizebox{\columnwidth}{!}{
  \begin{tabular}{lccc}
    \toprule[0.7pt]
    Risk level&Low risk&Medium risk&High risk\\
    \midrule
    Avg. stage var.&0.354 (0.003)&0.393 (0.003)&0.437 (0.005)\\
  \bottomrule[0.7pt]
\end{tabular}}
\end{table}

The results show that the health status of patients with high risk is more unstable, and patients with low risk have the most stable status. This is consistent with conclusions in medical researches that clinicians use physiologic stability index to evaluate patients' health risk (i.e. patients with unstable status have higher mortality risk)~\cite{yeh1984validation,halm1998time}.

\subsubsection{Case study}
\begin{figure}[ht]
\centering
\includegraphics[width=\columnwidth]{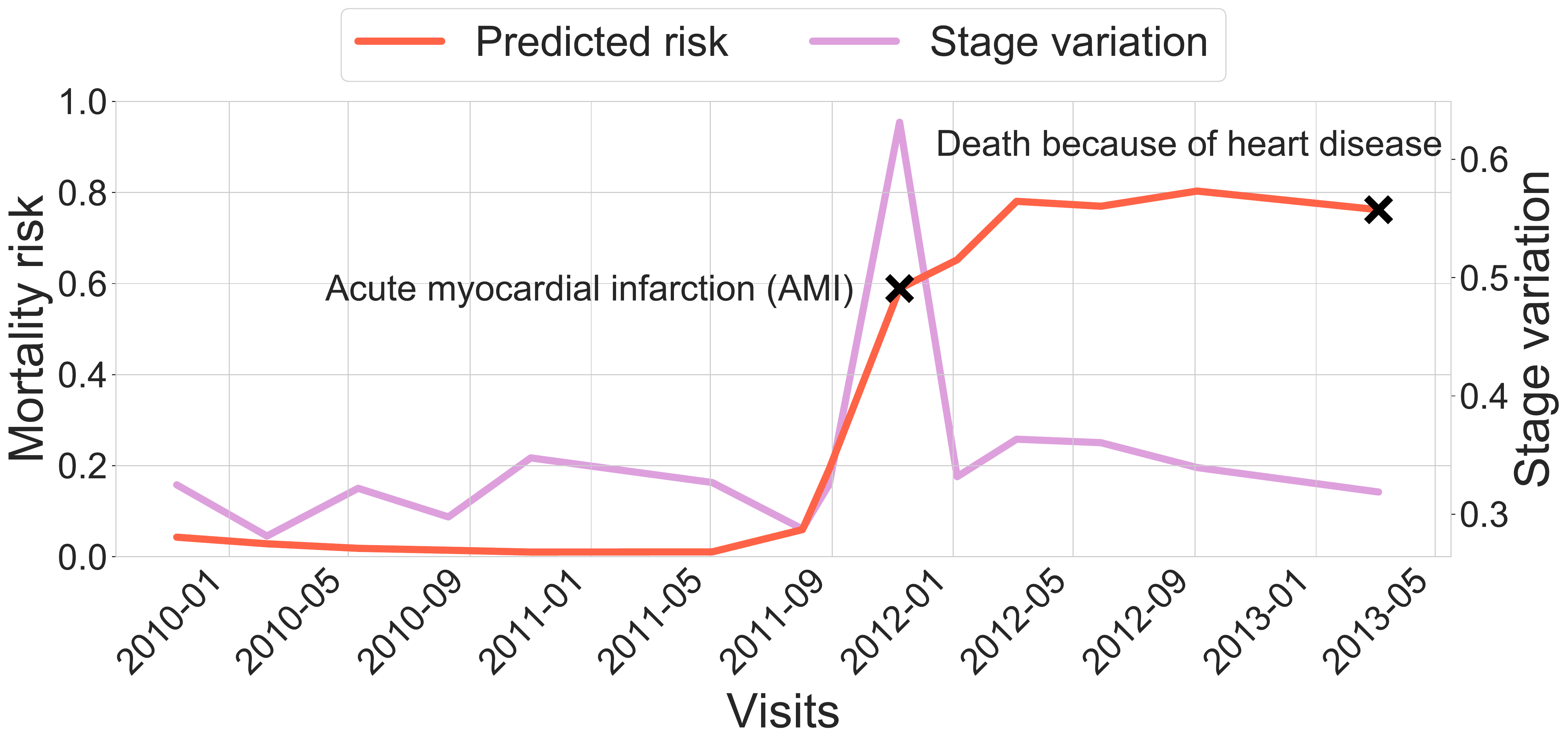}
\caption{Case study}
\label{fig:case}
\end{figure}

To explore how our model extracts the stage variation information of patients' health status and further utilize it to make predictions, we present a specific case study of a patient in the test set. As shown in Figure \ref{fig:case}, the purple line indicates the stage variation $s_{t}$ and the red line is the predicted mortality risk $y_{t}$ of the patient. 

For this specific patient, there is a distinct rising period in the two lines. Before May 2011, the patient's risk remains within a relatively low range. At the same time, $s_{t}$ also keeps a low value, which indicates that the model believes that the patient's health status is stable during this time. However, around 2012, the stage variation reaches the peak and the predicted risk rises rapidly, which means the patient has entered a high-risk stage. The risk remains a high value until the end. According to the clinical notes, the patient encountered Acute Myocardial Infarction (AMI) around 2012, and eventually died because of heart disease. If physicians were reminded when the model found the patient's status was changing drastically, the adverse outcome may be prevented or delayed by taking early interventions.

\subsection{Patient Subtyping}
In this experiment, we conduct patient subtyping on the ESRD dataset to investigate the expressive power of the patient representation learned from the \mname. ESRD is a chronic disease, and patients need to receive continuous medical treatment for years or even decades. Patients may face various risk factors such as infection, heart disease or cancer~\cite{weiner2004chronic}. Patient subtyping task is to seek patient groups with similar disease progression pathways~\cite{baytas2017patient}. Identifying patients subtypes can help clinicians develop targeted treatment plans and prevent adverse outcomes.

We use the learned $\widetilde{\bm{u}}_{t}$ at the last timestep in the previous risk prediction task as representations for patients' health status. For baseline models, we use the representations before the output layer. The learned representations are used to cluster the patients by the k-means algorithm~\cite{hartigan1979algorithm}. Since we do not know the ground truth groups of subtypes, we conduct several statistical analysis to assess the subtyping performance. Moreover, we use Calinski-Harabasz score~\cite{calinski1974dendrite} (C-H score for abbreviation) to evaluate the subtyping performance quantitatively. A higher C-H score relates to a model with better defined clusters. The C-H score is calculated as:
\begin{equation}
    \text{Calinski-Harabasz score} = \frac{tr(B_{k})}{tr_(W_{k})}\frac{m-k}{k-1}
\end{equation}
where $m$ is sample size, $k$ is the number of clusters, $B_{k}$ is the covariance matrix between clusters, $W_{k}$ is the covariance matrix within clusters, and $tr$ is the trace of matrix.

We fix a test set of 20\% of patients, and the other 80\% of patients are used for training. We use the same hyper-parameter as the risk prediction task. We tried several $k$ values for the k-means algorithm. We can observe four main clusters. Therefore we report the clustering Calinski-Harabasz score and average groud truth mortality risk in each cluster when $k=4$. The results are shown in Table~\ref{tab:cluster}.


\begin{table}[h!]
\centering
  \caption{Clustering results when $k=4$}
  \label{tab:cluster}
  \resizebox{1\columnwidth}{!}{
  \begin{tabular}{lccccc}
    \toprule[0.7pt]
    &C-H score&Cluster I&Cluster II&Cluster III&Cluster IV\\
    \midrule
    LSTM&74&0.09&0.19&0.33&0.59\\
    ON-LSTM&104&0.08&0.20&0.25&0.48\\
    T-LSTM&31&0.03&0.08&0.09&0.85\\
    Decay-LSTM&43&0.08&0.23&0.24&0.54\\
    $\text{Health-ATM}^{-}$&81&0.07&0.23&0.28&0.47\\
    \hline
    \mname&\textbf{165}&0.05&0.06&0.60&0.69\\
  \bottomrule[0.7pt]
\end{tabular}}
\end{table}

The results show \mname \  achieves over 58\% higher Calinski-Harabasz score compared to the best baseline model $\text{Health-ATM}^{-}$. The average risk in each cluster shows that \mname \  divide patients into two high-risk groups (Cluster III and IV) and two low-risk groups (Cluster I and II). On-LSTM, Decay-LSTM and $\text{Health-ATM}^{-}$ divide patients into one high-risk group (Cluster IV), two medium-risk groups (Cluster II and III) and one low-risk group (Cluster I). However, T-LSTM only identifies one high-risk group (Cluster IV) and three low-risk groups (Cluster I, II and III). Risk scores of different clusters for our baseline models are increasing evenly, which means the learned representations distribute evenly in latent space and not form meaningful clusters and thus results in low C-H score.

\subsubsection{High-risk patient subtypes}
In order to interpret the clustering results in terms of subtyping, we compared the medium-risk and high-risk clusters with low-risk clusters using T-test to identify discriminative features (p-value < 0.05). We find that there are 5-7 significant features in each cluster, and we report the top 5 significant features ranked by p-value in Table~\ref{tab:cluster_features}.

\begin{table}[h!]
\centering
  \caption{Most significant features (ranked by p-value) in medium-risk and high-risk clusters.}
  \label{tab:cluster_features}
  \resizebox{1\columnwidth}{!}{
  \begin{tabular}{lccc}
    \toprule[0.9pt]
    \multicolumn{4}{c}{\textbf{LSTM}}\\
    \midrule
    \multirow{2}{*}{Cluster II }&Albumin &C-rp&Blood chlorine\\
    &DBP&Glucose\\
    \midrule
    \multirow{2}{*}{Cluster III }&Albumin&C-rp&Blood chlorine\\
    &Glucose&DBP\\
    \midrule
     \multirow{2}{*}{Cluster IV}&Albumin&C-rp&DBP\\
     &Serum creatinine&Glucose\\
    \midrule [0.7pt]
    \multicolumn{4}{c}{\textbf{ON-LSTM}}\\
    \midrule
    \multirow{2}{*}{Cluster II }&Albumin&Glucose&C-rp\\
    &Blood urea&DBP\\
    \midrule
    \multirow{2}{*}{Cluster III}&Albumin&Glucose&DBP\\
    &Blood urea&C-rp\\
    \midrule
    \multirow{2}{*}{Cluster IV}&Albumin&DBP&C-rp\\
    &Glucose&Blood urea\\
    \midrule [0.7pt]
    \multicolumn{4}{c}{\textbf{T-LSTM}}\\
    \midrule
    \multirow{2}{*}{Cluster IV}&Albumin&Blood chlorine&Serum creatinine\\
    &Blood potassium&DBP\\
    \midrule [0.7pt]
    \multicolumn{4}{c}{\textbf{Decay-LSTM}}\\
    \midrule
    \multirow{2}{*}{Cluster II }&DBP&Albumin&Blood chlorine\\
    &hs-CRP&Blood sodium\\
    \midrule
    \multirow{2}{*}{Cluster III}&Albumin&DBP&Blood chlorine\\
    &Glucose&hs-CRP\\
    \midrule
    \multirow{2}{*}{Cluster IV}&DBP&Blood chlorine&Albumin\\
    &hs-CRP&Glucose\\
    \midrule [0.7pt]
    \multicolumn{4}{c}{\textbf{Health-ATM}}\\
    \midrule
    \multirow{2}{*}{Cluster II }&Blood chlorine&Blood potassium&Albumin\\
    &DBP&Blood urea\\
    \midrule
    \multirow{2}{*}{Cluster III}&Blood chlorine&Albumin&Blood potassium\\
    &DBP&Blood urea\\
    \midrule
    \multirow{2}{*}{Cluster IV}&Blood chlorine&Blood potassium&Albumin\\
    &DBP&Glucose\\
    \midrule [0.7pt]
    \multicolumn{4}{c}{\textbf{\mname}}\\
    \midrule
    \multirow{2}{*}{Cluster III}&Albumin&Serum creatinine&Blood urea\\
    &Appetite&C-rp\\
    \midrule
    \multirow{2}{*}{Cluster IV}&C-rp&DBP&Blood potassium\\
    &Albumin&Hemoglobin\\
  \bottomrule[0.9pt]
\end{tabular}}
\end{table}

The results show that almost all baseline models choose albumin, C-reactive protein, glucose, chlorine, diastolic blood pressure and blood urea to distinguish between low-risk patients and high-risk patients. However, significant features in different high-risk and medium-risk clusters are almost the same for all baseline models, which indicates that these models are unable to further distinguish subtypes among high-risk patients and therefore have worse clustering performance. 

In contrast, \mname \  clearly divides patients into two high-risk groups and also identifies more discriminative features. In Cluster III, albumin, blood urea and appetite are important indicators related to patients' nutritional status~\cite{gama2012serum,patel2013serum,di2017nutritional}. These biomarkers can reflect patients' health status from a long-term perspective. While in Cluster IV, blood potassium and diastolic blood pressure are important indicators for heart diseases such as heart failure~\cite{kannel1987heart,lee2009relation}. Patients with high white blood cell count and C-reactive protein are likely to have severe infections~\cite{jialal2001inflammation,franz1999comparison}. We also notice that \mname \  identifies hemoglobin as a significant feature in Cluster IV, which has never been identified by other baseline models. According to medical research, the constant reducing of hemoglobin is a key factor denoting the occurrence of acute GI bleeding \cite{Tomizawa2014Reduced}, which may cause sudden death.

\begin{figure}[h]
\centering
\includegraphics[width=0.95\columnwidth]{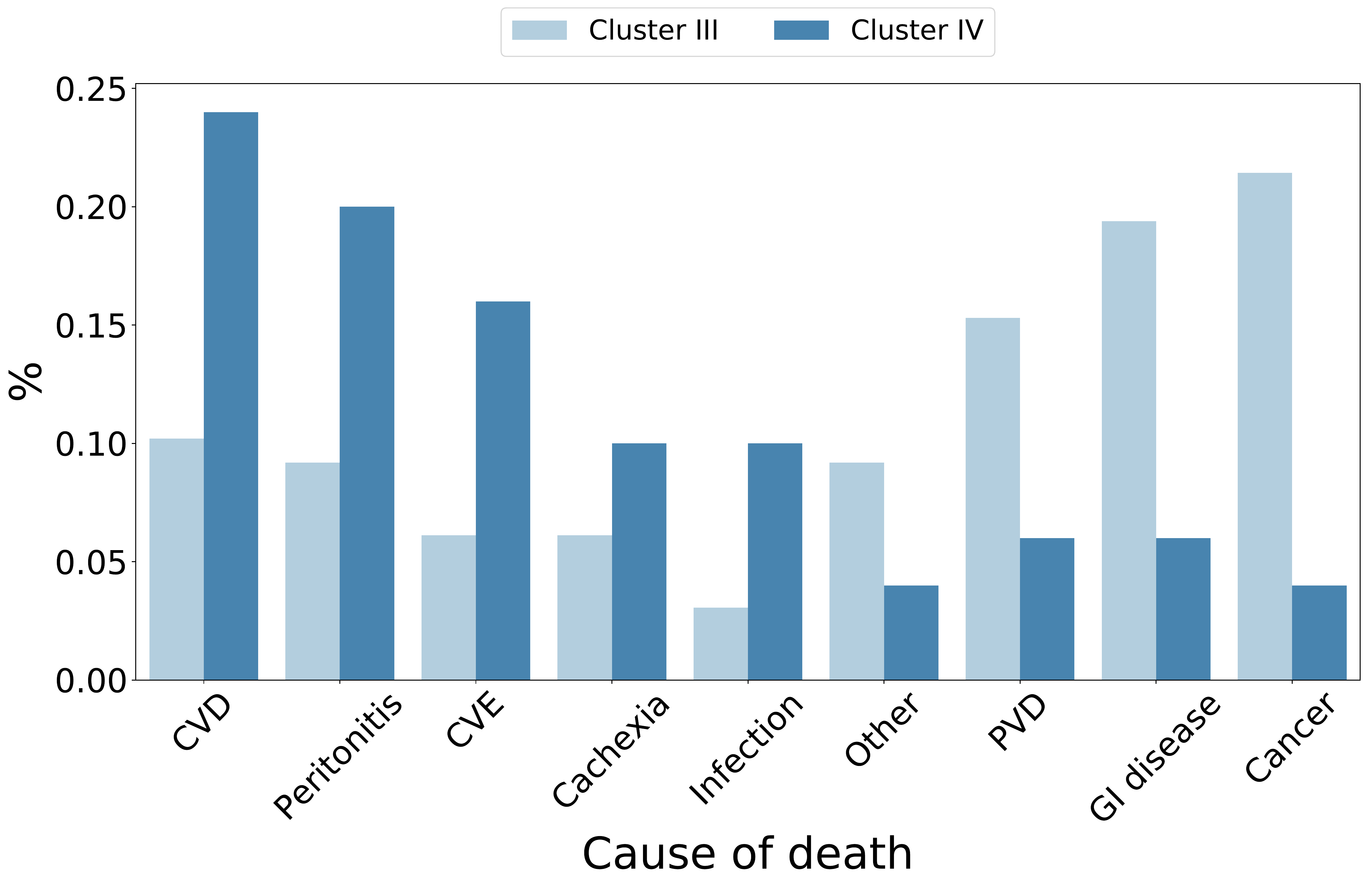}
\caption{Cause of death distribution for \mname \  results: The distribution of two clusters are very different, which leads to distinct phenotypes}
\label{fig:cluster_dist}
\end{figure}

\begin{figure}[h]
\centering
\includegraphics[width=0.95\columnwidth]{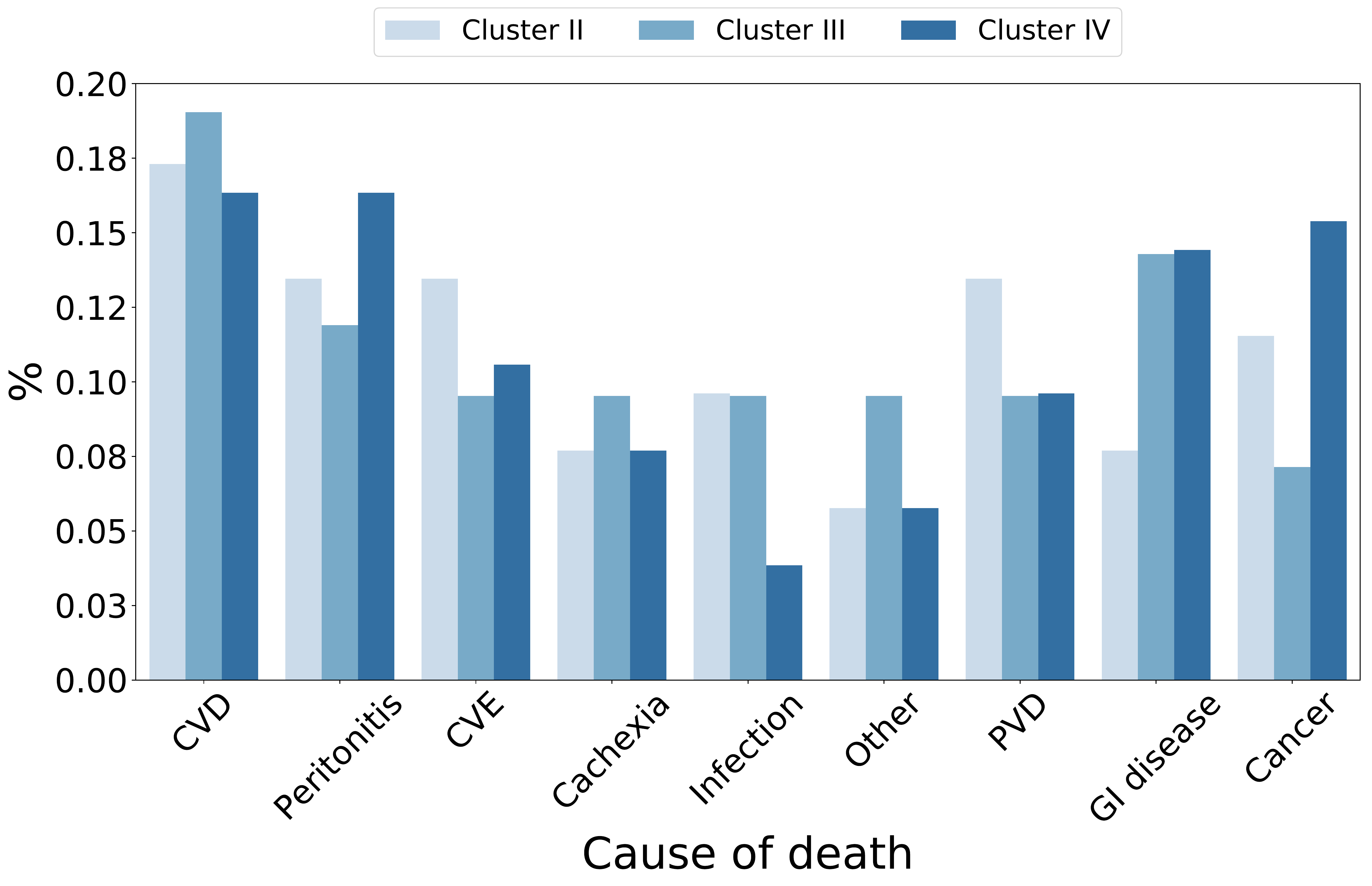}
\caption{Cause of death distribution for ON-LSTM results: Distribution of different clusters are very similar, which means lower quality of clusters}
\label{fig:cluster_dist2}
\vskip -1em
\end{figure}

We also explore the distribution of cause of death in Cluster III and Cluster IV of \mname, the result is shown in Fig~\ref{fig:cluster_dist}. In Cluster III, the main causes of death are cancer and gastrointestinal disease (GI disease), which are mainly considered as more chronic disease~\cite{derogatis1979psychological,prentice1978regression}. While in Cluster IV, the main causes of death are cardiovascular, peritonitis and cerebrovascular, which are acute symptoms~\cite{kannel1987heart,fried1996peritonitis,estanol1975cardiac}. This is consistent with medical researches and our previous experiment results. However, baseline models failed to identify these high-risk subtypes. For example, the cause of death distribution for ON-LSTM is shown in Fig~\ref{fig:cluster_dist2}. There is no significant difference in the cause of death distribution between different clusters.

\subsubsection{Low-risk patient subtypes}
In Table~\ref{tab:cluster}, \mname \  also divides low-risk patients into two subtypes (Cluster I and II). To identify the difference between the two clusters, we report the discriminative features in two clusters using T-test and the mean value of these features in each cluster. The result is shown in Table~\ref{tab:cluster_features2}.

\begin{table}[h!]
\centering
  \caption{Significant features (ranked by p-value) in Cluster I and Cluster II of \mname \  results. Even two low-risk clusters have quite different range of lab measures. }
  \label{tab:cluster_features2}
  \begin{tabular}{lcc}
    \toprule[0.7pt]
    Feature&Cluster I Mean&Cluster II Mean\\
    \midrule
    Serum creatinine&988.1  (172.4)&758.3 (112.8)\\
    Glucose&6.9  (1.4)&5.4 (1.0)\\
    Albumin&36.7  (3.2)&40.9 (2.5)\\
    Blood chlorine&100.1  (2.9)&97.5 (2.1)\\
  \bottomrule[0.7pt]
\end{tabular}
\end{table}

We can observe that patients in Cluster I have higher serum creatinine, glucose, blood chlorine and lower albumin compared to patients in Cluster II. As discussed above, albumin and serum creatinine are important indicators for patients' nutritional status, and indicate the severity of patients’ ESRD progression~\cite{di2017nutritional,patel2013serum}. Kidney damage may cause high blood chlorine level and high glucose level may indicate patients have diabetes~\cite{batlle1981pathogenesis}. In conclusion, patients in Cluster I have higher potential health risk compared to patients in Cluster II. 

\subsubsection{Health status stability vs. Patient subtypes}
In order to understand how patients' disease progression stage information help \mname \  to identify patient subtypes, we calculate the average stage variation of patients in each cluster. The results are shown in Table~\ref{tab:stage_subtype}

\begin{table}[h]
\centering
  \caption{Patient subtypes and status stability}
   
  \label{tab:stage_subtype}
  \resizebox{\columnwidth}{!}{
  \begin{tabular}{lcccc}
    \toprule[0.7pt]
    Cluster&I&II&III&IV\\
    \midrule
    Avg. stage var.&0.297 (0.044)&0.295 (0.039)&0.350 (0.038)&0.409 (0.031)\\
  \bottomrule[0.7pt]
\end{tabular}}
\end{table}

The results show that the variation of patients' health status is positively correlated with patients' health risk in each cluster, which have been proved in Table~\ref{tab:stage_risk}. We notice that patients in Cluster IV have the most unstable status, since they have more acute symptoms. Though patients in Cluster III still have high health risk, they have lower stage variation compared to Cluster IV, because their disease progressions are more chronic. The results in Table~\ref{tab:stage_subtype} are consistent with the observations and medical findings we discussed above. Compared to baseline models, \mname \  can learn discriminative patient representations from EHR sequences by extracting and utilizing the disease progression stage information.

\section{Conclusions}
In this work, we propose a stage-aware neural network model, \mname, to conduct health risk prediction using patients' stage variation of health status. \mname \  consists of a stage-aware LSTM module and a stage-adaptive convolutional module.  \mname \  can extract the stage of patients' health status at each visit unsupervisedly, then leverage and re-calibrate stage-related variation patterns into risk prediction. Supervised health risk prediction experiments on two real-world datasets demonstrate that \mname \  consistently outperforms state-of-the-art methods by better capturing inherent disease progression stage information in EHR data. Compared to the best baseline model, \mname \  achieves 10\% higher AUPRC and min(Re,P+) on public MIMIC-III dataset, and 12\% higher AURPC and 6\% higher min(Re,P+) on ESRD dataset. The patient subtyping experiment shows that \mname \  performs better than baseline models to learn discriminative representations by extracting and utilizing the stage information. In clinical practice, we hope our model can help physicians identify the patients with unstable health status to prevent or delay the adverse outcome.

\section{Acknowledgments}
This work is part supported by National Natural Science Foundation of China (No. 91546203), National Science Foundation award IIS-1418511, CCF-1533768 and IIS-1838042, the National Institute of Health award NIH R01 1R01NS107291-01 and R56HL138415.

\appendix
\section{Dataset details}

The basic statistics of two dataset are shown in Table~\ref{tab:statistics}.

\begin{table}[h]
\vskip -1em
\centering
  \caption{Basic Statistics of ESRD and MIMIC Dataset}
  \label{tab:statistics}
  \resizebox{1\columnwidth}{!}{
  \begin{tabular}{llc}
    \toprule[0.7pt]
    &Statistic&Value\\
    \midrule
    \multirow{7}{*}{ESRD}&\# patients & 656\\
    &\# patients died & 261\\
    &\# visit & 13091\\
    &\# visit with positive label ($y_{t}=1$) & 2287\\
    &\# visit with negative label ($y_{t}=0$)& 10804\\
    &\# average time interval between visits & 3.4 months \\
    &\% female & 49\% \\
    \hline
    \multirow{7}{*}{MIMIC-III}&\# patients & 33,678\\
    &\# ICU stays & 41,902\\
    &\# visit & 2,202,114\\
    &\# visit with positive label ($y_{t}=1$) & 45,364\\
    &\# visit with negative label ($y_{t}=0$)& 2,156,750\\
    &\% female & 44\% \\
  \bottomrule[0.7pt]
\end{tabular}}
\vskip -1em
\end{table}

\section{Implementation details}
For hyper-parameter settings of each baseline model, our principle is as follows: For some hyper-parameter, we will use the recommended setting if it is available in the original paper. Otherwise, we determine its value by grid search on the validation set.
 
\begin{itemize}
    \item \textbf{LSTM/T-LSTM/Decay-LSTM}. The hidden units of LSTM cell are set to 64 / 128 for ESRD / MIMIC-III dataset respectively.
    \item \textbf{ON-LSTM}. The hidden units of LSTM cell are set to 72 / 384. The chunk size factor $C$ is set to 36 / 128 for ESRD / MIMIC-III dataset respectively.
    \item $\textbf{Health-ATM}^{-}$. The hidden units of LSTM cell are set to 64 / 128. The number of convolutional filters is set to 32 / 64 and the size of filters is set to 3 / 5 for ESRD / MIMIC-III dataset respectively.
    \item \textbf{\mname}. The length of observation window $K$ is set to 10. The hidden units of LSTM cell are set to 72 / 384. The chunk size factor $C$ is set to 36 / 128.
\end{itemize}

Additionally, we use dropout \cite{srivastava2014dropout} before the output layer and dropconnect \cite{wan2013regularization} in the LSTM layer. The dropout rate is set to 0.5 / 0.3 for ESRD / MIMIC-III dataset respectively. We train each model for 50 epochs on MIMIC-III dataset and 200 epochs on the ESRD dataset. The learning rate is set to 0.001.

\newpage
\bibliographystyle{ACM-Reference-Format}
\bibliography{main}

\end{document}